\pdfoutput=1

\documentclass[11pt]{article}

\usepackage[final]{acl}

\usepackage{times}
\usepackage{latexsym}

\usepackage[T1]{fontenc}

\usepackage[utf8]{inputenc}

\usepackage{microtype}

\usepackage{inconsolata}

\usepackage{graphicx}

\usepackage{tabularx}
\usepackage{algorithm} 
\usepackage{algpseudocode} 
\usepackage{graphicx}
\usepackage{amsmath}
\usepackage{booktabs}
\usepackage{multirow}
\usepackage[export]{adjustbox}
\usepackage{makecell}

\title{PatentVision: A multimodal method for drafting patent applications}

  \author{Ruo Yang \\
  Samsung Semiconductor, Inc. \\
  San Jose, CA \\
  \texttt{r.yang@partner.samsung.com} \\\And
  Sai Krishna Reddy Mudhiganti \\
  Samsung Semiconductor, Inc. \\
  San Jose, CA \\
  \texttt{s.mudhiganti@samsung.com} \AND
  Manali Sharma  \\
  Samsung Semiconductor, Inc. \\
  San Jose, CA \\
  \texttt{manali.s@samsung.com}
  \\}

\begin{document}
\maketitle
\begin{abstract}
Patent drafting is complex due to its need for detailed technical descriptions, legal compliance, and visual elements. Although Large Vision-Language Models (LVLMs) show promise across various tasks, their application in automating patent writing remains underexplored. In this paper, we present PatentVision, a multimodal framework that integrates textual and visual inputs—such as patent claims and drawings—to generate complete patent specifications. Built on advanced LVLMs, PatentVision enhances accuracy by combining fine-tuned vision-language models with domain-specific training tailored to patents. Experiments reveal it surpasses text-only methods, producing outputs with greater fidelity and alignment with human-written standards. Its incorporation of visual data allows it to better represent intricate design features and functional connections, leading to richer and more precise results. This study underscores the value of multimodal techniques in patent automation, providing a scalable tool to reduce manual workloads and improve consistency. PatentVision not only advances patent drafting but also lays groundwork for broader use of LVLMs in specialized areas, potentially transforming intellectual property management and innovation processes.
\end{abstract}

\section{Introduction}

Drafting a comprehensive patent specification involves transforming intricate technical concepts, embodied in both written claims and accompanying illustrations, into precise and coherent legal documentation. Traditional methods predominantly focus on textual analysis, leveraging natural language processing techniques to interpret and generate patent specifications. However, these approaches often overlook the critical role of visual elements—patent drawings—which serve as indispensable carriers of design intent and functional details. As a result, existing systems struggle to fully capture the nuanced interplay between textual and visual components, leading to limitations in accurately reflecting inventors' intentions and meeting professional drafting standards.
\begin{figure}[!t]
    \centering
    \includegraphics[scale=.39]{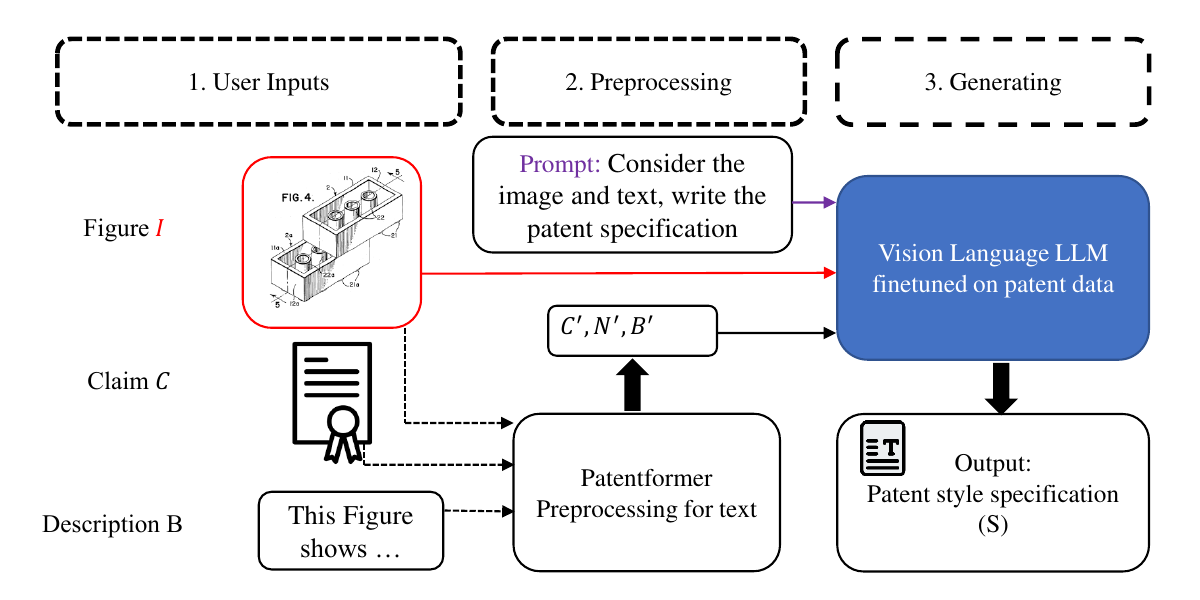}
    \vspace{-5mm}
    \caption{
    PatentVision is a framework that generates high-quality patent specifications using multimodal inputs like images, patent claims, and optional figure descriptions. Specifically, PatentVision integrates three inputs: the image, enriched textual content derived from PatentFormer’s text processing pipeline~\cite{PatentFormer}, and an instruction prompt tailored for the base vision-language model. The vision-language model is fine-tuned on domain-specific patent data to learn and replicate the formal writing style typical of patent specifications, thereby assisting patent authors in drafting coherent and contextually appropriate descriptions.
    }
    \label{fig:1}
    \vspace{-5mm}
\end{figure}
In recent years, advances in Large Vision-Language Models (LVLMs) have demonstrated significant potential in bridging the gap between linguistic and visual domains. By integrating multimodal data streams, LVLMs enable a deeper comprehension of contextually rich scenarios, offering new avenues for enhancing automated processes across diverse applications. This study investigates the application of state-of-the-art LVLMs, including models such as Gemma \cite{team2025gemma}, LLAVA \cite{liu2024llavanext}, and LLaMA \cite{grattafiori2024llama}, to address the challenges inherent in patent specification drafting. Specifically, we examine how these models can effectively combine patent claims and corresponding drawings to produce high-quality patent specifications. Through rigorous experimentation, our findings reveal that incorporating visual inputs significantly elevates the accuracy and coherence of generated texts, closely mirroring established human drafting practices.

The proposed framework employs a dual-input architecture, where textual inputs consist of patent claims and descriptive annotations, while visual inputs encompass detailed patent diagrams. By fusing these modalities, the system achieves a holistic interpretation of the invention, enabling it to generate specifications that are not only technically accurate but also aligned with legal requirements. These insights underscore the transformative potential of multimodal approaches in automating patent drafting, paving the way for more efficient and reliable intellectual property management.

\begin{figure*}[!h]
    \centering
    \includegraphics[scale=.35]{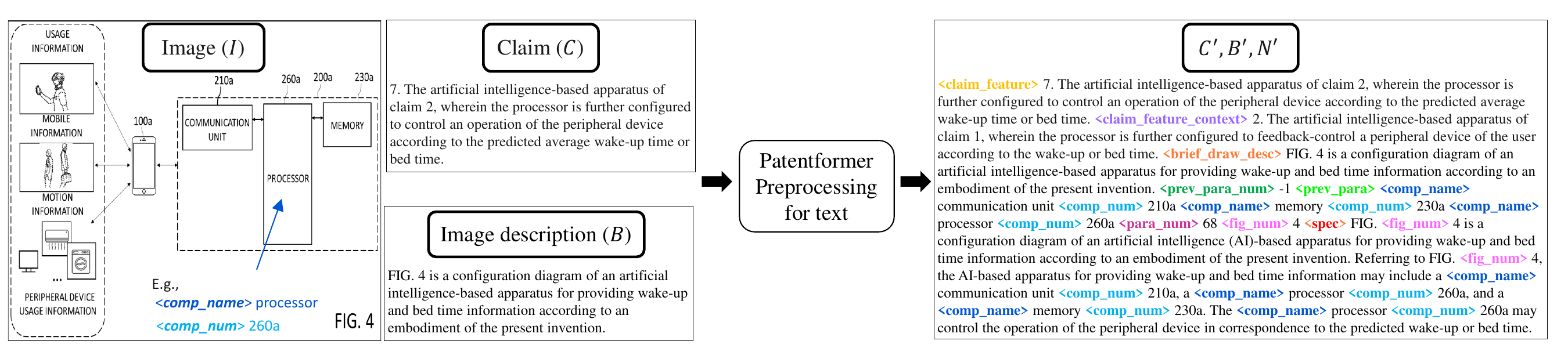}
    \vspace{-5mm}
    \caption{PatentFormer~\cite{PatentFormer} performs text processing by taking as input the image $I$, the claim $C$, and the image description $B$. It outputs an enriched textual representation containing structured tokens such as \texttt{$<$$comp_{\_}name$$>$}, which are subsequently encoded using the tokenizer of the language model. These enriched tokens provide explicit semantic anchors that facilitate more accurate and context-aware specification generation.}
    \label{fig:2}
    \vspace{-5mm}
\end{figure*}

\section{Related Work}
Most prior work on patent text generation has focused on specific sections rather than full specifications. For example, \citet{lee2020patent} fine-tuned GPT-2 to generate claims; \citet{lee2020patentper} added a BERT-based module for personalized claim generation; \citet{lee2019measuring} introduced a span-based framework for evaluating claim generation; and \citet{jiang2024patentclaims} generated claims from detailed descriptions. \citet{lee2020controlling} used structural metadata to control generation via text-to-text mappings, while \citet{lee2020measuringsem} applied semantic search for control. \citet{lee2023evaluating} pre-trained GPT-J on patent corpora for autocompletion and introduced the Autocomplete Effectiveness (AE) ratio, which \citet{jieh2022effectiveness} extended using bidirectional pretraining of GPT-J-6B. \citet{christofidellis2022pgt} proposed Patent Generative Transformer (PGT), a GPT-2-based model for part-specific generation. Other work focused on summarizing patents to produce titles \cite{souza2021comparative}, abstracts \cite{guoliang2023generating, zhu2023automatic}, prior art \cite{lee2020prior}, or figure captions \cite{aubakirova2023patfig}. Separately, research on modeling long documents in legal and medical domains has explored hierarchical transformers and efficient attention mechanisms, such as Longformer \cite{longformer2020}, Linformer \cite{linformer2020}, Big Bird \cite{bigbird2020}, and Hi-Transformer \cite{hitransformer2021}. BioGPT \cite{luo2022biogpt} fine-tuned GPT-2 for biomedical tasks. \citet{surveytextgeneration2024} provide a survey on pretrained language models for long-form generation.

In contrast to prior work focused on generating short sections or summaries, our approach builds on PatentFormer \cite{PatentFormer} and is, to our knowledge, the first work to generate full patent specifications directly from claims and drawings.

\section{Methodology}
\label{sec:methodology}
Formally, let $\mathcal{P}$ represent a patent document containing a sequence of $l$ claims, $\mathcal{C}$ = \{$c_1$, $c_2$, ..., $c_l$\}, a sequence of $m$ specification paragraphs, $\mathcal{S}$ = \{$s_1$, $s_2$, ..., $s_m$\}, a set of $t$ drawing images, $\mathcal{I}$ = \{$i_1$, $i_2$, ..., $i_t$\}, and a set of $t$ brief descriptions of the drawings, $\mathcal{B}$ = \{$b_1$, $b_2$, ..., $b_t$\}, corresponding to each image in $\mathcal{I}$. 
For $\forall i_z$ $\in$ $I$, let $n_z$ represent a set of $k$ pairs of component names and their respective component numbers that appear in the drawing; $n_z$ = \{<$i_{z_1}^{name}$, $i_{z_1}^{num}$>,<$i_{z_2}^{name}$, $i_{z_2}^{num}$>, ..., <$i_{z_k}^{name}$, $i_{z_k}^{num}$>\}, where $i_{z_j}^{name}$ is the name of $j^{th}$ component and $i_{z_j}^{num}$ is the number of $j^{th}$ component in image $i_z$; $\mathcal{N}$ = \{$n_1$, $n_2$, ..., $n_t$\} corresponding to all images in $\mathcal{I}$. Each image is preprocessed by first rotating it to the correct orientation and then rescaling it such that the maximum of its height or width is 4096 pixels. In a later section, we analyze the impact of image resolution on model performance and demonstrate that higher-resolution images lead to improved specification generation compared to lower-resolution settings.

\subsection{Claim+Diagram-to-Specification}
Instead of generating specifications from text input only in \cite{PatentFormer}, e.g., \textit{text-to-text}, we introduce a multimodal task (\textit{image-text-to-text}) called claim+diagram-to-specification, $\mathcal{T}$$\rightarrow$$\mathcal{S}$. Its goal is to generate output specification, $\mathcal{S}$, by using $\mathcal{C}$, $\mathcal{B}$, $\mathcal{I}$, and $\mathcal{N}$ as inputs, where the output specification must support all the input claim features, $\mathcal{C}$, correctly describe the drawings by using drawing descriptions, $\mathcal{B}$, the corresponding image associated with the claim, $\mathcal{I}$, and pairs of components, $\mathcal{N}$, associated with each drawing. 

We construct training samples containing the input and output pairs, <$\mathcal{T}, \mathcal{S}$>, where $\mathcal{T}$=<$\mathcal{C}, \mathcal{B}, \mathcal{I}, \mathcal{N}$>. Rather than learning from all the input text at once to produce the entire specification, we introduce an auxiliary task of mapping each claim feature to a paragraph in the specification and use only one drawing\footnote{Note that some paragraphs may describe more than one drawing. In this work, we assume that each paragraph describes only one drawing, and remove the lines from paragraph that refer to other figures.} associated with a paragraph. We first match $b_z$ to $s_y$ by checking for common figure numbers. Then, we match $s_y$ to $c_x$ by using the average of cosine similarity and BLEU scores between $s_y$ and $c_x$. Each $s_y \in \mathcal{S}$ may describe a figure or not. We only keep paragraphs that describe at least one figure in the patent by checking the presence of the words `FIG.', 'Fig.', and 'Figure', as well as occurrences of any component names and numbers in each paragraph. To simulate the extraction of component names and numbers from a drawing image $i_z$ in the training data, we extract $n_z$ from each $s_y$, as described in \cite{PatentFormer}.\footnote{USPTO provides patent drawings in .TIFF or .PDF formats, so the extraction of component names and numbers from images is not accurate; hence, we simulated the extraction of component names and numbers from specification, instead. In practice, the drawing files are usually provided in Visio or powerpoint formats, from which extracting the component names and numberings is straightforward.} 
Finally, we construct the quadruplets of samples, <$c_x$, $b_z$, $i_z$, $n_z$, $s_y$>, where <$c_x$, $b_z$, $i_z$, $n_z$> is the input to produce the corresponding output specification, $s_y$. We customize the tokenizer and insert special tags into the input and output tokens to help the model understand different contexts.

\subsection{PatentVision}
\label{sec:method_PatentFormer}
Now we introduce our multimodal model, PatentVision, that embeds rich context into the training data for generating specifications and uses patent images directly. Similar to \cite{PatentFormer}, first, for each claim feature extracted from an independent claim, we provide as context the remaining claims features of that claim, and for each claim feature extracted from a dependent claim, we provide as context any remaining claim features as well as its parent claim as context. Second, for each figure number, component name, and component number, we embed special tags in both the input and output specifications to mark their presence in the training data. Third, we also provide context by referencing the previous paragraph number and the current paragraph number to help the model understand the context and generate a coherent specification. As an real example shown in Figure~\ref{fig:2}, we represent the enriched versions of $\mathcal{C}$, $\mathcal{N}$, and $\mathcal{S}$ as $\mathcal{C'}$, $\mathcal{N'}$, $\mathcal{S'}$, and $\mathcal{B'}$=$\mathcal{B}$. 
As shown by \cite{PatentFormer}, embedding rich context into the training data yields significant improvements in the model's performance. 

Instead of relying solely on textual input, PatentVision integrates multimodal vision-language models to improve specification quality by incorporating both visual and textual information. PatentVision extends the capabilities of PatentFormer~\cite{PatentFormer} in two key aspects. First, it interprets and utilizes visual content from figures associated with patent claims to enhance the generation of specifications by jointly modeling visual and textual modalities. Second, unlike PatentFormer, which generates outputs solely based on enriched text inputs, PatentVision is designed as an interactive agent capable of engaging in dialogue with the human users. Specifically, it accepts human instructions, enriched textual descriptions, and visual inputs as part of the specification generation process. As a result, the generated specification can vary according to the provided human instructions, enabling greater flexibility.

\section{Experimental Setup}
In this section, we provide details of the experimental settings, including the dataset, models, baselines, evaluation metrics, and hardware specifics used for training and evaluation of PatentVision. 
    
    
    






\noindent \textbf{Dataset. } 
We construct the first dataset for generating specifications from the claims and associated drawings. We worked with four patent experts and focused on generating patents for a specific CPC code, `G06F'\footnote{\url{https://www.uspto.gov/web/patents/classification/cpc/html/defG06N.html\#G06F}} 
, which includes patents from a diverse range of topics related to electronic digital data processing. The dataset contains a total of 230K image-text-to-text samples. Due to the high computational cost of inference during evaluation, we randomly sample 1,000 instances as the test set, while the remaining samples are used for training.

\noindent \textbf{Models. } 
To train PatentVision, we evaluate three large vision-language models (LVLMs) as its core components: Gemma 3-12B \cite{team2025gemma}, LLAVA 1.6-13B \cite{liu2024llavanext}, and LLaMA 3.2-11B~\cite{grattafiori2024llama}. Each model is fine-tuned on the \texttt{Patent-2015-2023-G06F} dataset. Based on empirical performance, the best-performing model (Gemma 3) is selected for deployment within the PatentVision framework.
\\
\textbf{Baselines. } 
To the best of our knowledge, PatentFormer (T5-11B~\cite{JMLR:v21:20-074}) is the first work that addresses the task of generating specifications from both patent claims and corresponding drawings. As there is no prior baseline in the literature for direct comparison, we evaluate the performance of PatentVision against PatentFormer, the most closely related approach. For a fair comparison, we adopt the same post-processing strategy as described in~\cite{PatentFormer}, which ranks generated paragraphs based on alignment with input claims, component names, component numbers, and the correct figure number. The top-ranked paragraph is then selected as the final output.
\\
\textbf{Evaluation Metrics.}
To compare the models under various settings, we report the performance of PatentVision using ten popular metrics for natural language generation from the literature, including Bertscore~\cite{bert-score}, BLEU score~\cite{Papineni02bleu:a, lin-och-2004-orange}, ROUGE scores (R-1, R-2, R-L, and R-Lsum)~\cite{lin-2004-rouge}, WER~\cite{woodard1982}, Chrf~\cite{popovic-2017-chrf,popovic-2015-chrf}, METEOR~\cite{banarjee2005}, and NIST~\cite{10.5555/1289189.1289273}. 
 \\
\textbf{Training. } We utilized NVIDIA A100 GPUs (80 GB per GPU) for model training. Each model was trained for 1 epoch. Rather than fine-tuning the VL models directly, which requires a significant amount of GPU resources, we choose to fine-tune the models using LoRA \cite{lora} instead.

\section{Experimental Results}
\label{sec:results}

\begin{figure}[htb]
    \centering
    \includegraphics[scale=.25]{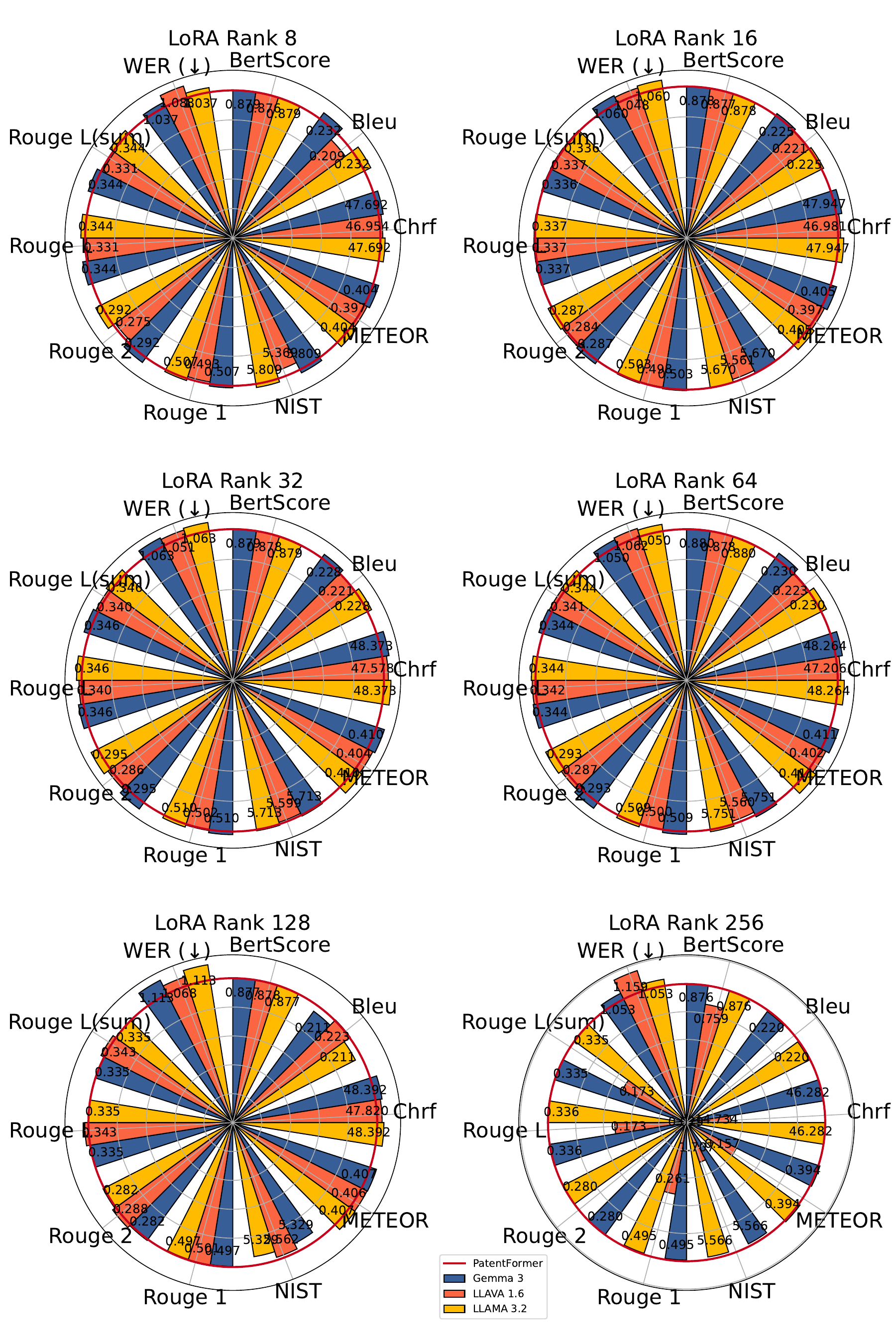}
    \vspace{-5mm}
    \caption{Comparison between PatentVision with different base LVLMs and LoRA ranks and PatentFormer.}
    \label{fig:vvsp}
    \vspace{-4mm}
\end{figure}

\begin{figure*}[htb]
    \centering
    \includegraphics[scale=.38]{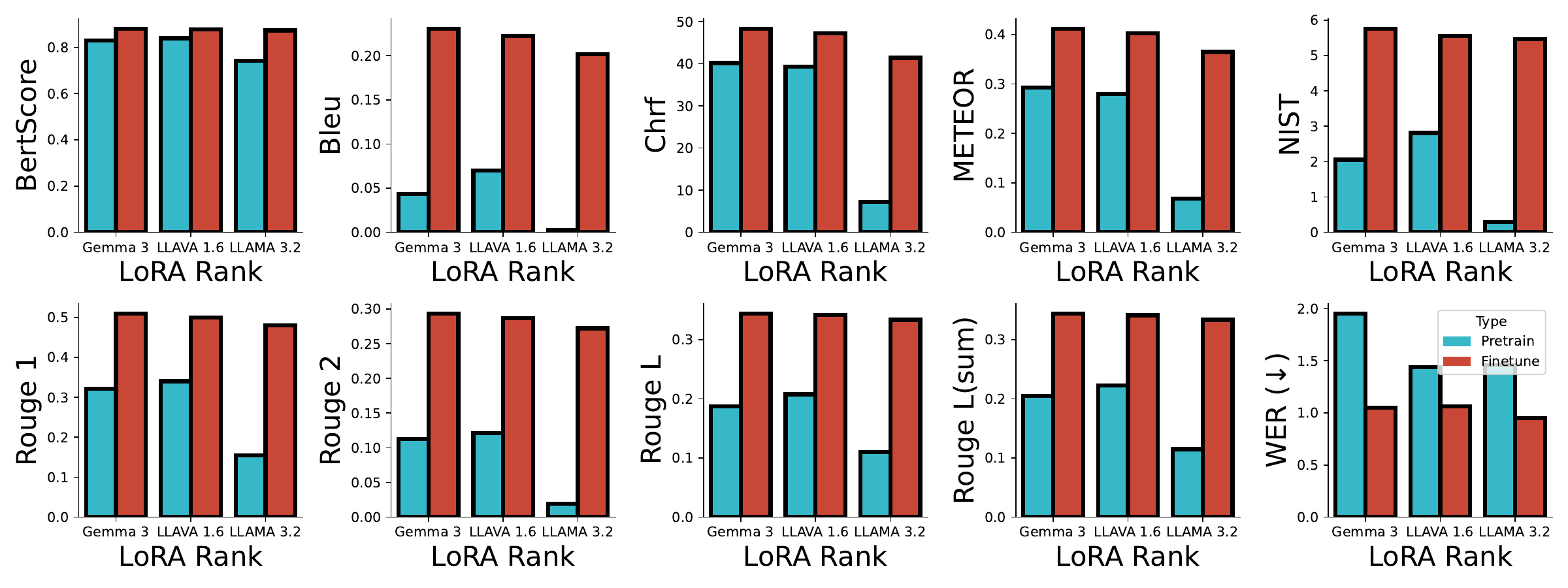}
    \vspace{-5mm}
    \caption{Performance of PatentVision with different base LVLMs compared to their pretrained versions. }
    \label{fig:pf}
    \vspace{-4mm}
\end{figure*}

\begin{figure*}[htb]
    \centering
    \includegraphics[scale=.4]{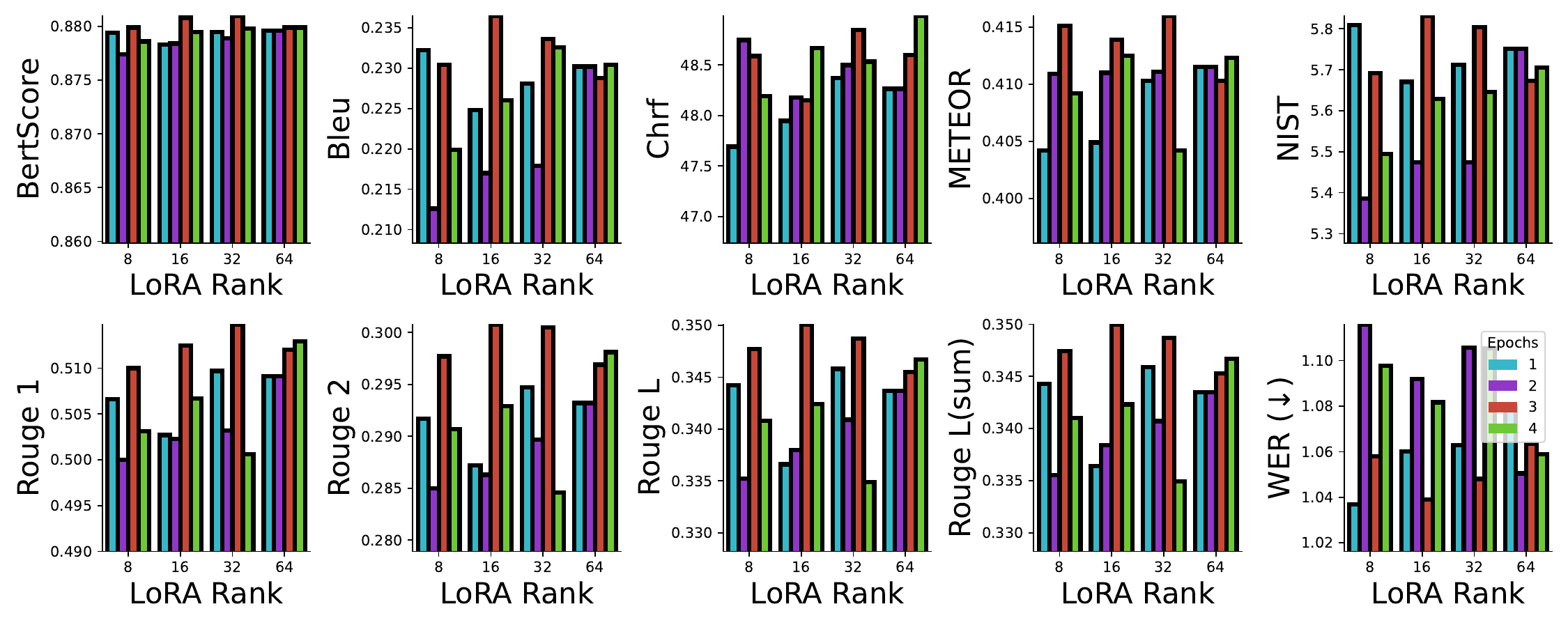}
    \vspace{-5mm}
    \caption{Performance of PatentVision with Gemma 3 as base model trained with varying epochs and LoRA ranks.}
    \label{fig:epoch}
    \vspace{-4mm}
\end{figure*}

\begin{figure*}[htb]
    \centering
    \includegraphics[scale=.4]{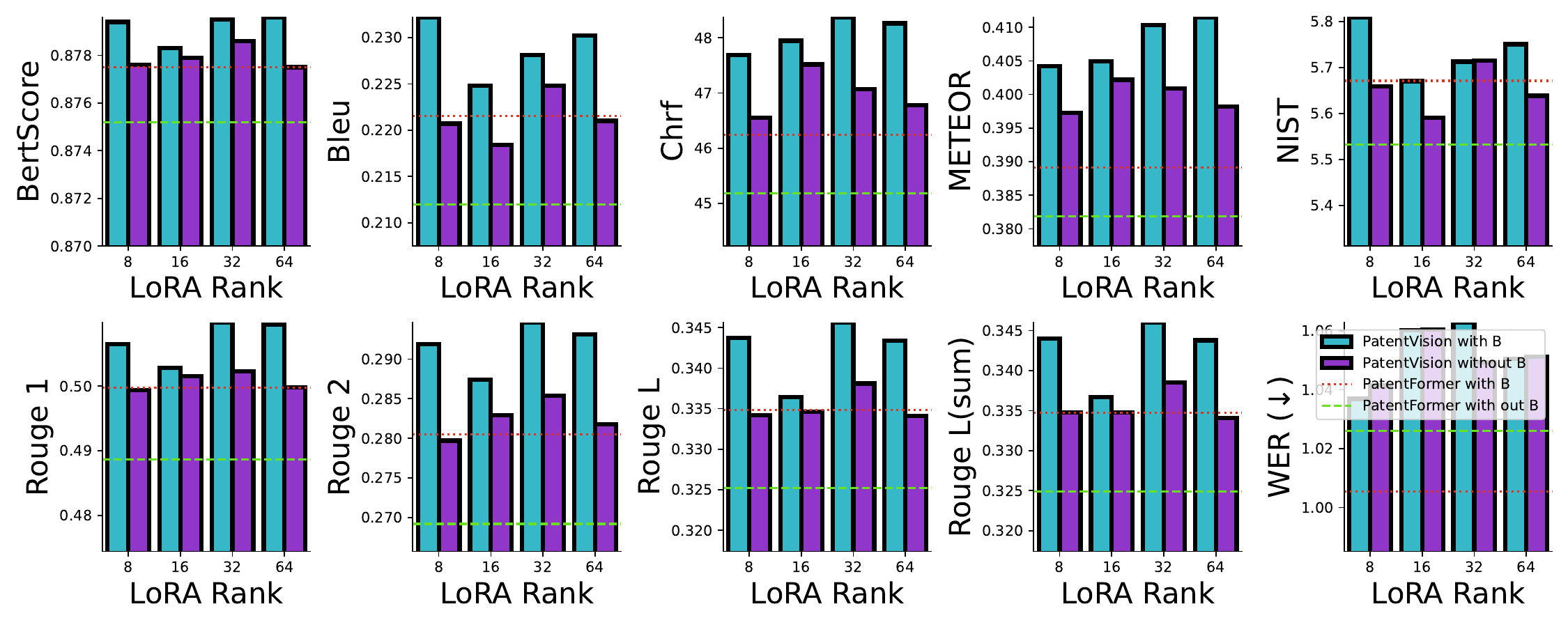}
    \vspace{-5mm}
    \caption{Performance of PatentVision with Gemma 3 as the base model trained with varying LoRA ranks on test sets with and without image descriptions ($B$).}
    \label{fig:noimage}
    \vspace{-4mm}
\end{figure*}

\begin{figure}[htb]
  \includegraphics[width=.45\textwidth]{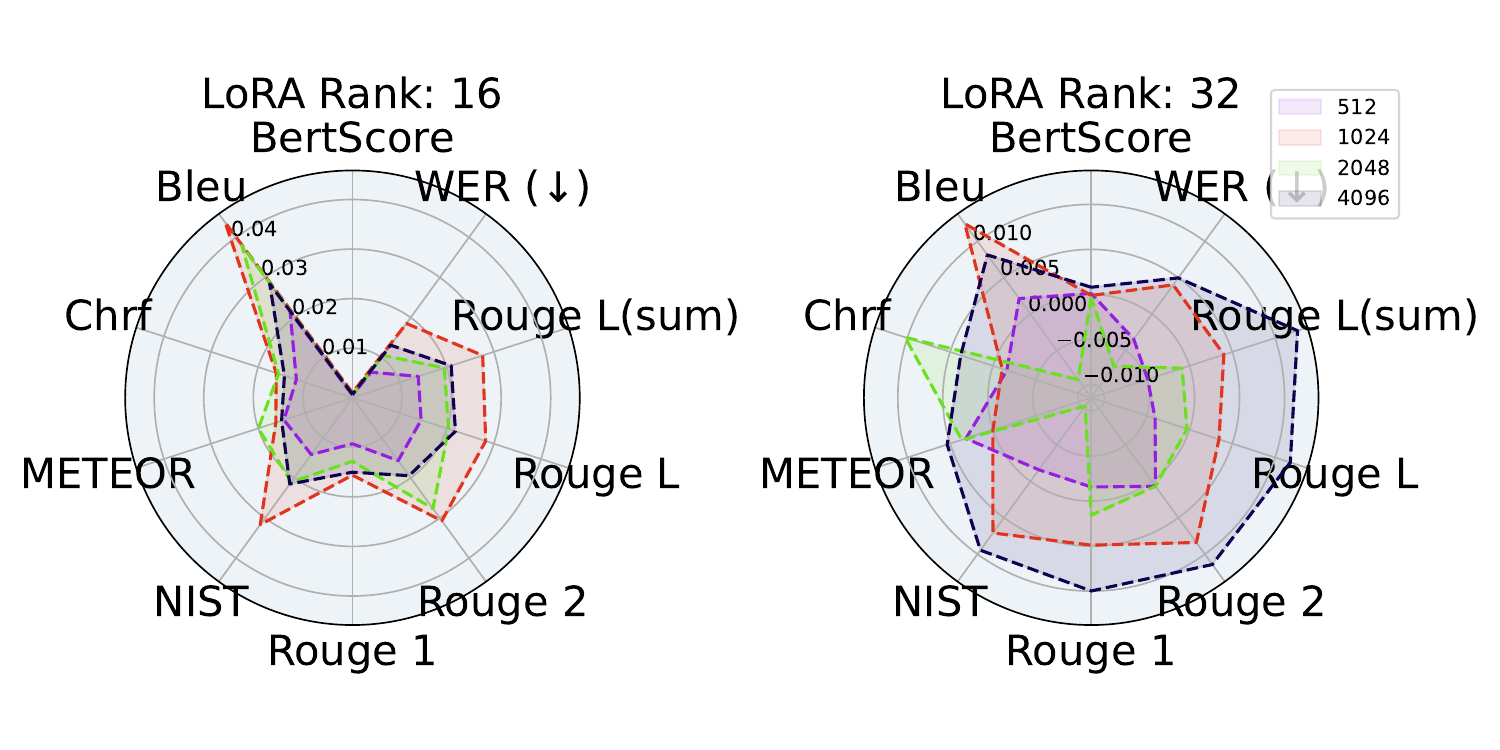}
  \vspace{-5mm}
  \caption{Performance improvement percentages compared to the performance of PatentVision using an image resolution of 256, for PatentVision with Gemma 3 as the base model across different metrics.}
  \label{fig:imagesize}
  \vspace{-4mm}
\end{figure}
In this section, we begin by comparing the performance of the multimodal PatentVision with the text-only PatentFormer to assess the benefits of incorporating visual understanding into the patent specification generation task. Next, we evaluate large vision-language models (LVLMs) on a dataset without image descriptions to demonstrate that PatentVision produces higher-quality outputs than PatentFormer, even when requiring less human input. We then compare fine-tuned LVLMs with their pretrained counterparts to quantify the quality improvements achieved through fine-tuning on our patent dataset. Finally, we examine the performance of LVLM across different training epochs and image resolutions to analyze the sensitivity of LVLMs to key hyperparameters. We additionally include evaluation tables in the appendix.

\subsection{PatentVision vs. PatentFormer}
To evaluate the benefits of incorporating visual understanding into the patent specification generation task, we first compare the performance of PatentVision, instantiated with different multimodal models, against the text-only PatentFormer. Specifically, we fine-tune PatentVision using Gemma 3, LLAVA 1.6, and LLaMA 3.2 as base models, each with varying LoRA ranks, on the Image-Text-to-Text patent data pairs. In parallel, PatentFormer is fine-tuned on the same dataset, but without access to image inputs. All models are trained for a single epoch to ensure consistent conditions. Figure~\ref{fig:vvsp} presents the results comparing PatentVision (with different base LVLMs) to PatentFormer. As shown, PatentVision consistently outperforms PatentFormer across all evaluation metrics.

\subsection{Finetuned vs. pretrained VL models}
Next, we evaluate the capability of original pretrained vision-language (VL) models on the patent specification generation task without any fine-tuning. This experiment allows us to assess the extent to which fine-tuning on our patent dataset improves model performance for domain-specific writing. Figure ~\ref{fig:pf} presents the results of both pretrained and fine-tuned VL models, where fine-tuning is performed with a LoRA rank of 64. The results clearly demonstrate that fine-tuned models substantially outperform their pretrained counterparts across all evaluation metrics, particularly in generating specifications consistent with legal and technical writing conventions.

\subsection{Removing image descriptions} Based on previous results, Gemma 3 outperforms both LLAVA 1.6 and LLaMA 3.2 on the patent specification generation task. Therefore, we focus subsequent analyses on PatentVision instantiated with Gemma 3 as the base model. We evaluate both PatentFormer and PatentVision (with Gemma 3) on the test dataset without any image descriptions , $B$. This setting allows us to assess whether PatentVision can learn to interpret visual content directly from raw images. As shown in Figure~\ref{fig:noimage}, 
as expected, removing the image description results in a slight performance degradation due to the reduced input information. However, PatentVision still significantly outperforms PatentFormer in the absence of image descriptions. Notably, PatentVision without image descriptions achieves better results than PatentFormer with image descriptions, demonstrating that PatentVision effectively extracts meaningful information directly from raw images.

\subsection{Ablation study}
Finally, we investigate the effects of varying LoRA ranks, image resolution, and number of epochs on the performance of PatentVision.
\\
\textbf{Impact of Lora Rank.} Using a small LoRA rank may limit the model’s capacity to acquire the domain-specific knowledge required for the patent writing task. Conversely, excessively large LoRA ranks can lead to convergence issues during training. Figure~\ref{fig:vvsp} shows PatentVision achieves better performance with mid-range LoRA ranks (e.g., 32, 64, and 128) across different base models. In contrast, training fails to converge with large LoRA ranks, e.g., 256 for LLAVA 1.6 and LLaMA 3.2.

\noindent \textbf{More epochs with Gemma 3. } As noted in the previous section, Gemma 3 outperforms LLAVA 1.6 and LLaMA 3.2 as the base model for PatentVision. To investigate the impact of training duration on generation quality, we train PatentVision using Gemma 3 across various LoRA ranks (8, 16, 32, and 64) with different numbers of training epochs. As shown in Figure~\ref{fig:epoch}, performance degrades when the model is trained for four epochs, indicating overfitting. In contrast, training for three epochs consistently yields superior results across different LoRA rank settings, indicating it as the optimal configuration for this task.

\noindent \textbf{The effects of image resolution. }
Next, we examine the effect of image resolution on the quality of the generated specifications. Specifically, we conduct experiments using image resolutions of 256, 512, 1024, 2048, and 4096 pixels. As shown in Figure~\ref{fig:imagesize}, higher image resolutions generally lead to improved generation quality. This trend suggests that increased resolution allows vision-language models to better capture and interpret fine-grained details within patent diagrams, which in turn enhances the overall specification generation.

\noindent \textbf{Chat functionality with Gemma 3. } One of the key advancements of PatentVision over PatentFormer is its design as an interactive agent capable of accepting human instructions, images, and patent text as input, rather than relying solely on patent text. This capability enables the users to provide post-generation instructions, such as editing or refining the generated specification, thereby supporting iterative improvement. The interactive nature of PatentVision significantly enhances the potential quality of the output, as the users can guide the model to correct or elaborate on its own generation—something not possible with PatentFormer. We plan to incorporate full conversational functionality in the next version of PatentVision to further support this interactive workflow.


\section{Conclusions}
We proposed a novel method, PatentVision, to utilize diverse patent-related information, e.g., patent claims, drawings, and brief descriptions of the drawings, for generating patent specification. We leveraged large vision language models to generate specification by using both text and image modalities. Experimental evaluations affirmed the effectiveness and practical usefulness of our proposed methods.

\clearpage
\section*{Ethics Statement}
Patents are legal documents, and the USPTO\footnote{https://www.federalregister.gov/documents/2024/04/11/2024-07629/guidance-on-use-of-artificial-intelligence-based-tools-in-practice-before-the-united-states-patent} recommends the practitioners to take extra care to verify the technical accuracy of the documents and compliance with 35 U.S.C. 112 when using AI drafting tools \cite{holman2024usptoAI}.

\bibliography{main}

\clearpage

\appendix

\end{document}